\documentclass[11pt,a4paper]{article}
\usepackage{array,multirow,graphicx}
\usepackage[hyperref]{emnlp-ijcnlp-2019}
\usepackage{times}
\usepackage{latexsym}
\usepackage{url}

\aclfinalcopy % Uncomment this line for the final submission

\title{Identifying Nuances in Fake News vs. Satire: \\Using Semantic and Linguistic Cues}

\author{Or Levi$^1$$^,$\thanks{\, Authors contributed equally} , Pedram Hosseini$^2$$^,$\footnotemark[1] , Mona Diab$^2$$^,$$^3$ \and David A. Broniatowski$^2$ \\
  $^1$AdVerifai \\
  $^2$The George Washington University, Washington DC, USA \\
  $^3$Amazon AWS AI, Seattle, USA \\
  {\tt or@adverifai.com, phosseini@gwu.edu}
  }

\date{}

\begin{document}
\maketitle
\begin{abstract}
The blurry line between nefarious fake news and protected-speech satire has been a notorious struggle for social media platforms. Further to the efforts of reducing exposure to misinformation on social media, purveyors of fake news have begun to masquerade as satire sites to avoid being demoted.
In this work, we address the challenge of automatically classifying fake news versus satire. Previous work have studied whether fake news and satire can be distinguished based on language differences.
Contrary to fake news, satire stories are usually humorous and carry some political or social message. We hypothesize that these nuances could be identified using semantic and linguistic cues. Consequently, we train a machine learning method using semantic representation, with a state-of-the-art contextual language model, and with linguistic features based on textual coherence metrics. 
Empirical evaluation attests to the merits of our approach compared to the language-based baseline and sheds light on the nuances between fake news and satire.
As avenues for future work, we consider studying additional linguistic features related to the humor aspect, and enriching the data with current news events, to help identify a political or social message.
\end{abstract}

\section{Introduction}

The efforts by social media platforms to reduce the exposure of users to misinformation have resulted, on several occasions, in flagging legitimate satire stories. To avoid penalizing publishers of satire, which is a protected form of speech, the platforms have begun to add more nuance to their flagging systems. Facebook, for instance, added an option to mark content items as ``Satire'', if ``the content is posted by a page or domain that is a known satire publication, or a reasonable person would understand the content to be irony or humor with a social message'' \cite{facebook}. This notion of humor and social message is also echoed in the definition of satire by Oxford dictionary as ``the use of humour, irony, exaggeration, or ridicule to expose and criticize people's stupidity or vices, particularly in the context of contemporary politics and other topical issues''. 

The distinction between fake news and satire carries implications with regard to the exposure of content on social media platforms. While fake news stories are algorithmically suppressed in the news feed,
the satire label does not decrease the reach of such posts. This also has an effect on the experience of users and publishers. For users, incorrectly classifying satire as fake news may deprive them from desirable entertainment content, while identifying a fake news story as legitimate satire may expose them to misinformation. For publishers, the distribution of a story has an impact on their ability to monetize content.

Moreover, in response to these efforts to demote misinformation, fake news purveyors have begun to masquerade as legitimate satire sites, for instance, carrying small badges at the footer of each page denoting the content as satire \cite{golbeck_fake}. The disclaimers are usually small such that the stories are still being spread as though they were real news \cite{poynter}.

This gives rise to the challenge of classifying fake news versus satire based on the content of a story. While previous work \cite{golbeck_fake} have shown that satire and fake news can be distinguished with a word-based classification approach, our work is focused on the semantic and linguistic properties of the content. Inspired by the distinctive aspects of satire with regard to humor and social message, our hypothesis is that using semantic and linguistic cues can help to capture these nuances. 

Our main research questions are therefore, RQ1) are there semantic and linguistic differences between fake news and satire stories that can help to tell them apart?; and RQ2) can these semantic and linguistic differences contribute to the understanding of nuances between fake news and satire beyond differences in the language being used?

The rest of paper is organized as follows: in section \ref{sec:related}, we briefly review studies on fake news and satire articles which are the most relevant to our work. In section \ref{sec:methods}, we present the methods we use to investigate semantic and linguistic differences between fake and satire articles. Next, we evaluate these methods and share insights on nuances between fake news and satire in section \ref{sec:evaluation}. Finally, we conclude the paper in section \ref{sec:conclusion} and outline next steps and future work.

\section{Related Work}
\label{sec:related}
Previous work addressed the challenge of identifying fake news \cite{Conroy_automatic,shu2017fake}, or identifying satire \cite{Burfoot_automatic,Reganti_modeling,Rubin_fake}, in isolation, compared to real news stories. 

The most relevant work to ours is that of Golbeck et al. \cite{golbeck_fake}. They introduced a dataset of fake news and satirical articles, which we also employ in this work. The dataset includes the full text of 283 fake news stories and 203 satirical stories, posted between January 2016 and October 2017, with a main focus on American politics. These fake and satirical stories were verified manually such that each fake news article is paired with a rebutting article from a reliable source. This data carries two desirable properties. First, the labeling is based on the content and not the source, and stories spread across a diverse set of sources. Second, as also mentioned in \cite{golbeck_fake}, the fact that fake news and satire articles both focus on American politics minimizes the possibility that the topic of the articles will influence the classification.

In their work, Golbeck et al. studied whether there are differences in the language of fake news and satirical articles on the same topic that could be utilized with a word-based classification approach. A model using the Naive Bayes Multinomial algorithm is proposed in their paper which serves as the baseline in our experiments.

%they point out a paper that concluded "we found that the headlines were especially relevant to detecting satire"

%maybe we can use bert to calculate language model probability of the sentences as a feature (similar to semantic relatedness)

% https://arxiv.org/pdf/1902.11145.pdf
% data is not based on sources, but based on content
% german dataset https://www.ims.uni-stuttgart.de/forschung/ressourcen/korpora/germansatire.html
% only word embeddings

% https://www.sciencedirect.com/science/article/abs/pii/S0950705117301855
% https://www.sciencedirect.com/science/article/abs/pii/S0950705116305226
% https://ieeexplore.ieee.org/tt/document/7079363

\section{Method}
\label{sec:methods}
In the following subsections, we investigate the semantic and linguistic differences of satire and fake news articles.\footnote{Reproducibility report, including codes and results, is available at: \href{https://github.com/adverifai/Satire_vs_Fake}{https://github.com/adverifai/Satire\_vs\_Fake}}

\subsection{Semantic Representation with BERT}
To study the semantic nuances between fake news and satire, we use BERT \cite{devlin2018bert}, which stands for Bidirectional Encoder Representations from Transformers, and represents a state-of-the-art contextual language model. BERT is a method for pre-training language representations, meaning that it is pre-trained on a large text corpus and then used for downstream NLP tasks.
Word2Vec \cite{Mikolov_distributed} showed that we can use vectors to properly represent words in a way that captures semantic or meaning-related relationships. While Word2Vec is a context-free model that generates a single word-embedding for each word in the vocabulary, BERT generates a representation of each word that is based on the other words in the sentence. It was built upon recent work in pre-training contextual representations, such as ELMo \cite{Peters_deep} and ULMFit \cite{Howard_universal}, and is deeply bidirectional, representing each word using both its left and right context. We use the pre-trained models of BERT and fine-tune it on the dataset of fake news and satire articles using Adam optimizer with 3 types of decay and 0.01 decay rate. Our BERT-based binary classifier is created by adding a single new layer in BERT's neural network architecture that will be trained to fine-tune BERT to our task of classifying fake news and satire articles.

\smallskip
\begin{table*}[t]
\begin{tabular}{|lclcccl|}
%\cline{2-6}
\hline
                                           & PCA Component                  & Description                               & estimate & std.error & statistic & \textbf{} \\ %\cline{1-6}
                                           \hline
\multicolumn{1}{|l|}{\multirow{16}{*}{\rotatebox[origin=c]{90}{Satire associated}}} & \textbf{RC19} & First person singular pronoun incidence   & 1.80     & 0.41      & 4.38      & ***                        \\
\multicolumn{1}{|l|}{}                                      & \textbf{RC5}  & Sentence length, number of words          & 0.66     & 0.18      & 3.68      & ***                        \\
\multicolumn{1}{|l|}{}                                      & \textbf{RC15} & Estimates of hypernymy for nouns          & 0.61     & 0.19      & 3.18      & **                         \\
\multicolumn{1}{|l|}{}                                      & \textbf{RC49} & Word Concreteness                         & 0.54     & 0.17      & 3.18      & **                         \\
\multicolumn{1}{|l|}{}                                      & \textbf{RC35} & Ratio of casual particles to causal verbs & 0.56     & 0.18      & 3.10      & **                         \\
\multicolumn{1}{|l|}{}                                      & \textbf{RC91} & Text Easability PC Referential cohesion   & 0.45     & 0.16      & 2.89      & **                         \\
\multicolumn{1}{|l|}{}                                      & \textbf{RC20} & Incidence score of gerunds                & 0.43     & 0.16      & 2.77      & **                         \\
\multicolumn{1}{|l|}{}                                      & RC32          & Expanded temporal connectives incidence   & 0.44     & 0.16      & 2.75      & **                         \\
\multicolumn{1}{|l|}{}                                      & \textbf{RC9}  & Third person singular pronoun incidence   & 0.44     & 0.16      & 2.67      & **                         \\
\multicolumn{1}{|l|}{}                                      & RC43          & Word length, number of letters            & 0.45     & 0.20      & 2.27      & *                          \\
\multicolumn{1}{|l|}{}                                      & RC46          & Verb phrase density                       & 0.37     & 0.16      & 2.25      & *                          \\
\multicolumn{1}{|l|}{}                                      & \textbf{RC97} & Coh-Metrix L2 Readability                 & 0.34     & 0.16      & 2.16      & *                          \\
\multicolumn{1}{|l|}{}                                      & \textbf{RC61} & Average word frequency for all words      & 0.50     & 0.24      & 2.13      & *                          \\
\multicolumn{1}{|l|}{}                                      & RC84          & The average givenness of each sentence    & 0.37     & 0.18      & 2.11      & *                          \\
\multicolumn{1}{|l|}{}                                      & RC65          & Text Easability PC Syntactic simplicity   & 0.38     & 0.18      & 2.08      & *                          \\
\multicolumn{1}{|l|}{}                                      & RC50          & Lexical diversity                         & 0.37     & 0.18      & 2.05      & *                          \\ \hline
\multicolumn{1}{|l|}{\multirow{9}{*}{\rotatebox[origin=c]{90}{Fake news associated}}} & \textbf{RC30} & Agentless passive voice density           & -1.05    & 0.21      & -4.96     & ***                        \\
\multicolumn{1}{|l|}{}                                      & \textbf{RC73} & Average word frequency for content words  & -0.72    & 0.20      & -3.68     & ***                        \\
\multicolumn{1}{|l|}{}                                      & \textbf{RC59} & Adverb incidence                          & -0.62    & 0.18      & -3.43     & ***                        \\
\multicolumn{1}{|l|}{}                                      & \textbf{RC55} & Number of sentences                       & -0.79    & 0.26      & -3.09     & **                         \\
\multicolumn{1}{|l|}{}                                      & RC62          & Causal and intentional connectives        & -0.42    & 0.15      & -2.72     & **                         \\
\multicolumn{1}{|l|}{}                                      & \textbf{RC34} & LSA overlap between verbs                 & -0.35    & 0.16      & -2.22     & *                          \\
\multicolumn{1}{|l|}{}                                      & \textbf{RC44} & LSA overlap, adjacent sentences           & -0.36    & 0.16      & -2.16     & *                          \\
\multicolumn{1}{|l|}{}                                      & RC47          & Sentence length, number of words          & -0.36    & 0.18      & -2.03     & *                          \\
\multicolumn{1}{|l|}{}                                      & RC89          & LSA overlap, all sentences in paragraph   & -0.34    & 0.17      & -1.97     & *                          \\
\multicolumn{1}{|l|}{}                                      & (Intercept)   &                                           & -0.54    & 0.19      & -2.91     &                            \\ %\cline{1-6}
\hline
\end{tabular}
\caption{Significant components of our logistic regression model using the Coh-Metrix features. Variables are also separated by their association with either satire or fake news. \textbf{Bold}: the remaining features following the step-wise backward elimination. \textit{Note}: *** p $<$ 0.001, **  p $<$ 0.01, *  p $<$ 0.05.}\label{tab:cohmetrix-results}
\end{table*}

\subsection{Linguistic Analysis with Coh-Metrix}
Inspired by previous work on satire detection, and specifically Rubin et al. \cite{Rubin_fake} who studied the humor and absurdity aspects of satire by comparing the final sentence of a story to the first one, and to the rest of the story - we hypothesize that metrics of text coherence will be useful to capture similar aspects of semantic relatedness between different sentences of a story.

Consequently, we use the set of text coherence metrics as implemented by Coh-Metrix \cite{mcnamara2010coh}. Coh-Metrix is a tool for producing linguistic and discourse representations of a text. As a result of applying the Coh-Metrix to the input documents, we have 108 indices related to text statistics, such as the number of words and sentences; referential cohesion, which refers to overlap in content words between sentences; various text readability formulas; different types of connective words and more. To account for multicollinearity among the different features, we first run a Principal Component Analysis (PCA) on the set of Coh-Metrix indices. Note that we do not apply dimensionality reduction, such that the features still correspond to the Coh-Metrix indices and are thus explainable. Then, we use the PCA scores as independent variables in a logistic regression model with the fake and satire labels as our dependent variable. Significant features of the logistic regression model are shown in Table \ref{tab:cohmetrix-results} with the respective significance levels. We also run a step-wise backward elimination regression. Those components that are also significant in the step-wise model appear in bold.

\section{Evaluation}
\label{sec:evaluation}
In the following sub sections, we evaluate our classification model and share insights on the nuances between fake news and satire, while addressing our two research questions.

\subsection{Classification Between Fake News and Satire}
We evaluate the performance of our method based on the dataset of fake news and satire articles and using the F1 score with a ten-fold cross-validation as in the baseline work \cite{golbeck_fake}.

First, we consider the semantic representation with BERT.
Our experiments included multiple pre-trained models of BERT with different sizes and cases sensitivity, among which the large uncased model, \textbf{bert\_uncased\_L-24\_H-1024\_A-16}, gave the best results. We use the recommended settings of hyper-parameters in BERT's Github repository and use the fake news and satire data to fine-tune the model. Furthermore, we tested separate models based on the headline and body text of a story, and in combination. Results are shown in Table \ref{tab:bert-results}. The models based on the headline and text body give a similar F1 score. However, while the headline model performs poorly on precision, perhaps due to the short text, the model based on the text body performs poorly on recall. The model based on the full text of headline and body gives the best performance.

To investigate the predictive power of the linguistic cues, we use those Coh-Metrix indices that were significant in both the logistic and step-wise backward elimination regression models, and train a classifier on fake news and satire articles. We tested a few classification models, including Naive Bayes, Support Vector Machine (SVM), logistic regression, and gradient boosting - among which the SVM classifier gave the best results.

Table \ref{tab:clf-results} provides a summary of the results. We compare the results of our methods of the pre-trained BERT, using both the headline and text body, and the Coh-Mertix approach, to the language-based baseline with Multinomial Naive Bayes from \cite{golbeck_fake}\footnote{We were not able to reproduce the same result as in the original paper, most possibly due to the difference in the toolkits used. Hence, to make our results comparable, we replicated the experiments of the original paper using the toolkits we used in our experiments.}. Both the semantic cues with BERT and the linguistic cues with Coh-Metrix significantly outperform the baseline on the F1 score. The two-tailed paired t-test with a 0.05 significance level was used for testing statistical significance of performance differences. The best result is given by the BERT model. Overall, these results provide an answer to research question RQ1 regarding the existence of semantic and linguistic difference between fake news and satire.

\begin{table}[t]
\begin{tabular}{|c|c|c|c|}
\hline
\textbf{Model} & \textbf{P} & \textbf{R} & \textbf{F1} \\ \hline
Headline only           & 0.46               & 0.89            & 0.61              \\ \hline
Text body only          & 0.78               & 0.52            & 0.62              \\ \hline
Headline + text body    & \textbf{0.81}               & \textbf{0.75}            & \textbf{0.78}     \\ \hline
\end{tabular}
\caption{Results of classification between fake news and satire articles using BERT pre-trained models, based on the headline, body and full text. \textbf{Bold}: best performing model. \textit{P}: Precision, and \textit{R}: Recall}\label{tab:bert-results}
\end{table}

%\begin{table}[t]
%\begin{tabular}{|l|c|}
%\hline
%\textbf{Method}           & \textbf{F1} \\ \hline
%Baseline: Multinomial Naive Bayes                  & 0.67        \\ \hline
%Logistic Regression - Coh-Metrix & 0.73*        \\ \hline
%Pre-trained BERT - full text           & 0.78*        \\ \hline
%\end{tabular}
%\caption{Summary of results of classification between fake news and satire articles using the baseline Multinomial Naive Bayes method, the linguistic cues of text coherence and semantic representation with a pre-trained BERT model. Statistically significant differences with the baseline are marked with '*'.}\label{tab:clf-results}
%\end{table}

\begin{table}[]
\begin{tabular}{|l|c|c|c|}
\hline
\textbf{Method}          & \textbf{P}                & \textbf{R}                & \textbf{F1}               \\ \hline
Baseline & 0.70                      & 0.64                      & 0.67                      \\ \hline
Coh-Metrix               & 0.72                       & 0.66                       & 0.74*                      \\ \hline
Pre-trained BERT         & \textbf{0.81} & \textbf{0.75} & \textbf{0.78}* \\ \hline
\end{tabular}
\caption{Summary of results of classification between fake news and satire articles using the baseline Multinomial Naive Bayes method, the linguistic cues of text coherence and semantic representation with a pre-trained BERT model. Statistically significant differences with the baseline are marked with '*'. \textbf{Bold}: best performing model. \textit{P}: Precision, and \textit{R}: Recall. For Coh-Metrix, we report the \textit{mean} Precision, Recall, and F1 on the test set.}\label{tab:clf-results}
\end{table}

\subsection{Insights on Linguistic Nuances}
With regard to research question RQ2 on the understanding of semantic and linguistic nuances between fake news and satire - a key advantage of studying the coherence metrics is explainability. While the pre-trained model of BERT gives the best result, it is not easily interpretable. The coherence metrics allow us to study the differences between fake news and satire in a straightforward manner.

Observing the significant features, in bold in Table \ref{tab:cohmetrix-results}, we see a combination of surface level related features, such as sentence length and average word frequency, as well as semantic features including LSA (Latent Semantic Analysis) overlaps between verbs and between adjacent sentences. 
%Presence of semantic features, which are associated with the bottom-line meaning of documents, among significant predictors to some degree supports our hypothesis about distinguishing fake from satire news articles.
Semantic features which are associated with the gist representation of content are particularly interesting to see among the predictors since based on Fuzzy-trace theory \cite{reyna2012new}, a well-known theory of decision making under risk, gist representation of content drives individual's decision to spread misinformation online. Also among the significant features, we observe the causal connectives, that are proven to be important in text comprehension, and two indices related to the text easability and readability, both suggesting that satire articles are more sophisticated, or less easy to read, than fake news articles.

\section{Conclusion and Future Work}
\label{sec:conclusion}
We addressed the challenge of identifying nuances between fake news and satire. Inspired by the humor and social message aspects of satire articles, we tested two classification approaches based on a state-of-the-art contextual language model, and linguistic features of textual coherence. Evaluation of our methods pointed to the existence of semantic and linguistic differences between fake news and satire. In particular, both methods achieved a significantly better performance than the baseline language-based method. Lastly, we studied the feature importance of our linguistic-based method to help shed light on the nuances between fake news and satire. For instance, we observed that satire articles are more sophisticated, or less easy to read, than fake news articles.

Overall, our contributions, with the improved classification accuracy and towards the understanding of nuances between fake news and satire, carry great implications with regard to the delicate balance of fighting misinformation while protecting free speech. 

For future work, we plan to study additional linguistic cues, and specifically humor related features, such as absurdity and incongruity, which were shown to be good indicators of satire in previous work. Another interesting line of research would be to investigate techniques of identifying whether a story carries a political or social message, for example, by comparing it with timely news information.

\bibliography{emnlp-ijcnlp-2019}
\bibliographystyle{acl_natbib}

\end{document}